\documentclass[sigconf]{acmart}
\usepackage{popets}
\usepackage{subcaption}
\usepackage{makecell}
\usepackage{amsmath}

\setcopyright{none}
\copyrightyear{}

\acmYear{}
\acmVolume{}
\acmNumber{}
\acmDOI{}
\acmISBN{}
\renewcommand\footnotetextcopyrightpermission[1]{}

\begin{document}

\title[TabularARGN]{Privacy-Preserving Tabular Synthetic Data Generation Using TabularARGN}

%%%%%%%%%%%%%%%% Authors' Info %%%%%%%%%%%%%%%%%

\author{Andrey Sidorenko}
\authornote{Both authors contributed equally to this research.}
\affiliation{%
  \institution{MOSTLY AI}
  \city{Vienna}
  \country{Austria}}
\email{andrey.sidorenko@mostly.ai}

\author{Paul Tiwald}
\authornotemark[1]
\affiliation{%
  \institution{MOSTLY AI}
  \city{Vienna}
  \country{Austria}}
\email{paul.tiwald@mostly.ai}


\renewcommand{\shortauthors}{Sidorenko et al.}

\begin{abstract}
Synthetic data generation has become essential for securely sharing and analyzing sensitive data sets. Traditional anonymization techniques, however, often fail to adequately preserve privacy. We introduce the Tabular Auto-Regressive Generative Network (TabularARGN), a neural network architecture specifically designed for generating high-quality synthetic tabular data. Using a discretization-based auto-regressive approach, TabularARGN achieves high data fidelity while remaining computationally efficient. We evaluate TabularARGN against existing synthetic data generation methods, showing competitive results in statistical similarity, machine learning utility, and detection robustness. We further perform an in-depth privacy evaluation using systematic membership-inference attacks, highlighting the robustness and effective privacy-utility balance of our approach.
\end{abstract}

% \keywords{data sets, neural networks, gaze detection, text tagging}

\maketitle
\pagestyle{plain} 

\section{Introduction}
Equitable and broad access to data is essential for advancing research, driving innovation, and addressing pressing societal challenges \cite{hradec2022multipurpose}. However, privacy concerns and the well‑documented shortcomings of traditional anonymisation techniques frequently result in valuable public- and private-sector data remaining inaccessible \cite{gadotti2024anonymization}. Safely unlocking this data can yield substantial benefits for organisations and society. Consider, for instance, a healthcare dataset containing sensitive patient information such as medical diagnoses, treatments, and demographic details. Disclosing this data poses significant privacy risks, potentially leading to patient re-identification and breaches of confidentiality.
Consequently, sensitive healthcare datasets often remain inaccessible, limiting research into disease patterns, treatment effectiveness, and healthcare optimisation. Synthetic data presents a promising solution, providing a secure means to share and analyse sensitive datasets by preserving their analytical utility while mitigating privacy risks \cite{drechsler2011synthetic, jordon2022synthetic, united2023synthetic, hu2023sokprivacypreservingdatasynthesis,challagundla2025synthetictabulardatageneration,10.1145/3711896.3736562}.

Synthetic data tools fundamentally act as density estimators, learning the underlying joint probability distributions of the original data. By leveraging deep neural networks to approximate these patterns and relationships from original data sets, synthetic samples can be generated that are structurally consistent, statistically representative, and genuinely novel, thereby effectively minimizing disclosure risks. The benefits of synthetic data extend beyond privacy protection, enabling arbitrary data volume generation, rebalancing of underrepresented groups, imputation of missing values, and conditional, scenario-specific sampling \cite{NEURIPS2021_ba9fab00, vanbreugel2023privacynavigatingopportunitieschallenges}.

In this paper, we introduce the Tabular Auto-Regressive Generative Network (TabularARGN), a neural network framework specifically designed for synthetic tabular data generation. TabularARGN addresses the density estimation challenge through an auto-regressive approach, approximating the joint distribution of tabular data as a sequence of conditional probabilities. Departing from approaches that treat tabular data rows as sequences of text \cite{hegselmann2023tabllm, smolyak2024large, miletic2024}, TabularARGN explicitly exploits structured value ranges and discretization, achieving high-fidelity synthetic data generation while maintaining simplicity and computational efficiency.

Beyond introducing TabularARGN, we conduct a rigorous analysis of privacy protection mechanisms within synthetic data generation methods, including an evaluation through Membership-Inference Attacks (MIAs) \cite{hayes2018loganmembershipinferenceattacks,7958568, pilgram2025consensusprivacymetricsframework, hyeong2022empiricalstudymembershipinference, wu2024inferenceattackstaxonomysurvey,vanbreugel2023membershipinferenceattackssynthetic}. While prior benchmarks often relied on metrics like Distance to Closest Record (DCR) \cite{zha24-TabSyn, mueller2025continuousdiffusionmixedtypetabular}, recent studies have highlighted limitations in these metrics \cite{yao2025dcrdelusionmeasuringprivacy}. Our comprehensive privacy analysis addresses this gap, providing deeper insights into the privacy robustness of TabularARGN-generated data and emphasizing the necessity of adopting stronger privacy evaluation methodologies for synthetic data generation studies. We also provide open-source access to TabularARGN to facilitate reproducibility and encourage practical adoption in the privacy research community.

\section{Related Work}
\label{sec:related_work}

To the best of our knowledge, \citet{uri16} were the first to develop an auto-regressive neural network (NADE) for tackling the problem of unsupervised distribution and density estimation. This foundational idea and architecture serve as the backbone of the TabularARGN flat and sequential models.

\citet{uri16} calculate the log-likelihood performance of NADE on binary tabular and binary image data sets, as well as purely real-valued data sets. However, they do not extend their work to generate synthetic data samples. Furthermore, the NADE architecture is not adapted in their work to handle multi-categorical data sets with variables of cardinality $\geq 2$, nor mixed-type data sets that combine categorical and numeric variables, both of which are critical for real-world tabular data synthesis tasks. 
NADE shares similarities with Bayesian Networks (BNs), as both methods factorize the joint probability into conditional distributions. 
However, BNs represent dependencies between variables using a directed acyclic graph. 
Once the conditional probability distributions are learned, data can be sampled from the model, making BNs a natural choice for generating tabular data \cite{ank15, qia23}.

Variational Autoencoders (VAEs), initially proposed by Kingma and Welling (2013) \cite{kingma13-vae}, and Generative Adversarial Networks (GANs), introduced by Goodfellow et al.\ (2014)  \cite{goo14-gan}, represent two prominent classes of generative models extensively used for synthetic data generation. Both methodologies have undergone significant adaptations specifically targeting mixed-type tabular data. For example, VAE-based approaches have been successfully employed in various studies, demonstrating robustness and efficiency in handling mixed-type data \cite{xu19, akr22-VAE, liu23-goggle}. Similarly, GAN-based frameworks have also been widely adapted and applied, showcasing their effectiveness across diverse domains, including healthcare data synthesis and electronic health record generation \cite{par18, xu19, zha21, qia23, li23-ehrMGAN, zha24}. These adaptations underscore the versatility and ongoing evolution of generative models, continually expanding their applicability and enhancing their performance in producing high-quality synthetic tabular data sets.

With the rise in popularity of Large Language Models (LLMs), token-based transformers have also been applied to generate tabular synthetic data based on a specific target data set. Some methods fine-tune pre-trained LLMs directly, while others train LLM-like architectures from scratch \cite{bor23, sol23-Realtabformer, kar24synehrgy}. 

In addition to token-based transformers, auto-regressive transformer models specifically designed for tabular data have also been proposed \cite{led21, cas23, gul23-TabMT}. Unlike LLMs or token-based approaches, these models explicitly leverage the inherent structure of tabular data, such as limited value ranges, column-specific distributions, and inter-feature relationships, to enhance both the efficiency and quality of synthetic data generation. More recently, researchers have begun exploring hybrid approaches that combine transformers with diffusion to model discrete features \cite{zha24-TabDar}. 

Transformer-based foundation models specifically tailored for tabular data have recently gained traction, notably TabPFN \cite{hollmann23tabpfn, hollmann25tabpfn}, a prominent example of a transformer-based in-context learner. Although primarily developed for rapid prediction on tabular machine learning tasks, these models inherently capture complex distributions and can also be adapted for synthetic tabular data generation \cite{ma24tabpfgen}.

Diffusion models, originally developed for image and audio synthesis, have also been adapted to generate tabular synthetic data \cite{lee23-Codi, kim23-StaSy, kot23-TabDDPM, zha24-TabSyn, vil24, mueller2025continuousdiffusionmixedtypetabular}. These models leverage iterative denoising processes to approximate complex data distributions and have shown promise in capturing intricate patterns in tabular data sets. Recent work shows that diffusion models can effectively memorize training data with a larger number of training epochs \cite{fang2024understandingmitigatingmemorizationdiffusion}.

There is an ongoing discussion in the machine learning community about why tree-based methods often outperform neural networks when applied to tasks involving tabular data, such as prediction \cite{gri22-NNvsTree}. This observation has inspired the development of tree-based generative models for tabular data \cite{jol24-ForrestDiffusion, watson23-ARF, mcc24-UnmaskingTrees}.

Given this context of generative modeling techniques, a key aspect of synthetic data generation that demands attention, especially within privacy-sensitive applications, is the incorporation of privacy-preserving mechanisms such as differential privacy (DP). Existing DP-based approaches for synthetic data generation broadly fall into two categories: deep generative models incorporating DP directly into training \cite{xie18dpgan, yoon19pategan, cas23} and methods synthesizing data based on differentially private queries of marginal distributions \cite{vie22, mckenna2022aim}. While the former leverage deep learning architectures for expressive modeling, the latter provide greater interpretability by explicitly constructing distributions from low-dimensional marginals.

This work makes the following contributions:
(i) we introduce TabularARGN, a simple yet effective neural architecture tailored for synthetic tabular data; (ii) we provide a tested and maintained reference implementation of TabularARGN under a fully permissive Apache v2 Open Source license\footnote{\url{https://github.com/mostly-ai/mostlyai-engine/}};
(iii) we rigorously evaluate TabularARGN against state-of-the-art methods, reporting its competitive performance across statistical similarity, machine learning utility, and detection robustness; (iv) we perform a detailed privacy evaluation, employing systematic MIAs to assess TabularARGN's robust protection mechanisms incorporated within a synthetic tabular data generation method.

\section{TabularARGN: Methodology and Features}
In this section, we introduce the flat model of the TabularARGN framework, a neural network architecture specifically designed to generate tabular synthetic data. The flat model synthesizes "flat tables", i.e., tabular data sets, where rows represent independent and identically distributed (i.i.d.) records. It utilizes a relatively simple and shallow feed-forward design, implementing an ("any-order") auto-regressive framework across discretized features, i.e., columns, of a data set. 

Before training TabularARGN, all columns in the original data set must be discretized. Categorical columns naturally fulfill this requirement. Numerical columns are discretized either by binning values into percentiles or by splitting values into individual digits, where each digit position forms a categorical sub-column. Date-time columns are decomposed into discrete components, such as year, month, day, and time. For geospatial data, latitude and longitude values are discretized by mapping them into categorical quadtiles, with the resolution automatically adjusted according to the data density of a given region.

As a result, each column from the original data set, irrespective of its initial data type, is transformed into one or more discrete sub-columns. Throughout the remainder of this paper, $D$ denotes the number of these discrete sub-columns. After sampling, this discretization mapping is reversed to convert the synthetic data back into its original format.

\subsection{The Flat Table Model}
Through the auto-regressive approach, TabularARGN splits the joint probability distribution of the data into a product of conditional probabilities $p(\mathbf{x}) = \prod_{i=1}^{D} p(x_i \mid x_{<i})$, where $x<i$ is the set of features preceding $x_i$. For each feature $x_i$, the model learns the discrete conditional probability $\hat{p}(x_i \mid x_{<i})$.

TabularARGN can be trained either in the "fixed-order" setting, where the column order is fixed and defined by the input data set or the user, or the "any-order" setting, where the order of features is dynamically shuffled for each training batch. This allows the model to estimate not only "fixed-order" conditional probabilities $\hat{p}(x_i \mid x_{<i})$, but also probabilities conditioned on any subset of features:
\begin{equation}
\quad \hat{p}(x_i \mid \{x_j : j \in S\}), \forall S \subseteq \{1,\dots,D\} \setminus \{i\}.
\end{equation}
This flexibility ensures that the model can adapt to arbitrary auto-regressive conditioning scenarios during generation, which is essential for applications such as imputation, fairness adjustments, and flexible conditional generation.

A similar order-agnostic approach was already suggested by \citet{uri16} within the NADE framework, adapted to the natural language processing domain \cite{yan19-xlnet,shih2022}, and is also discussed in more recent implementations of flat-table synthetic data generation \cite{led21, gul23-TabMT, mcc24-UnmaskingTrees}.

\subsection{Model Components and Information Flow} 
\begin{figure*}
     \centering
     \begin{subfigure}[b]{0.45\textwidth}
         \centering
         \includegraphics[width=\textwidth]{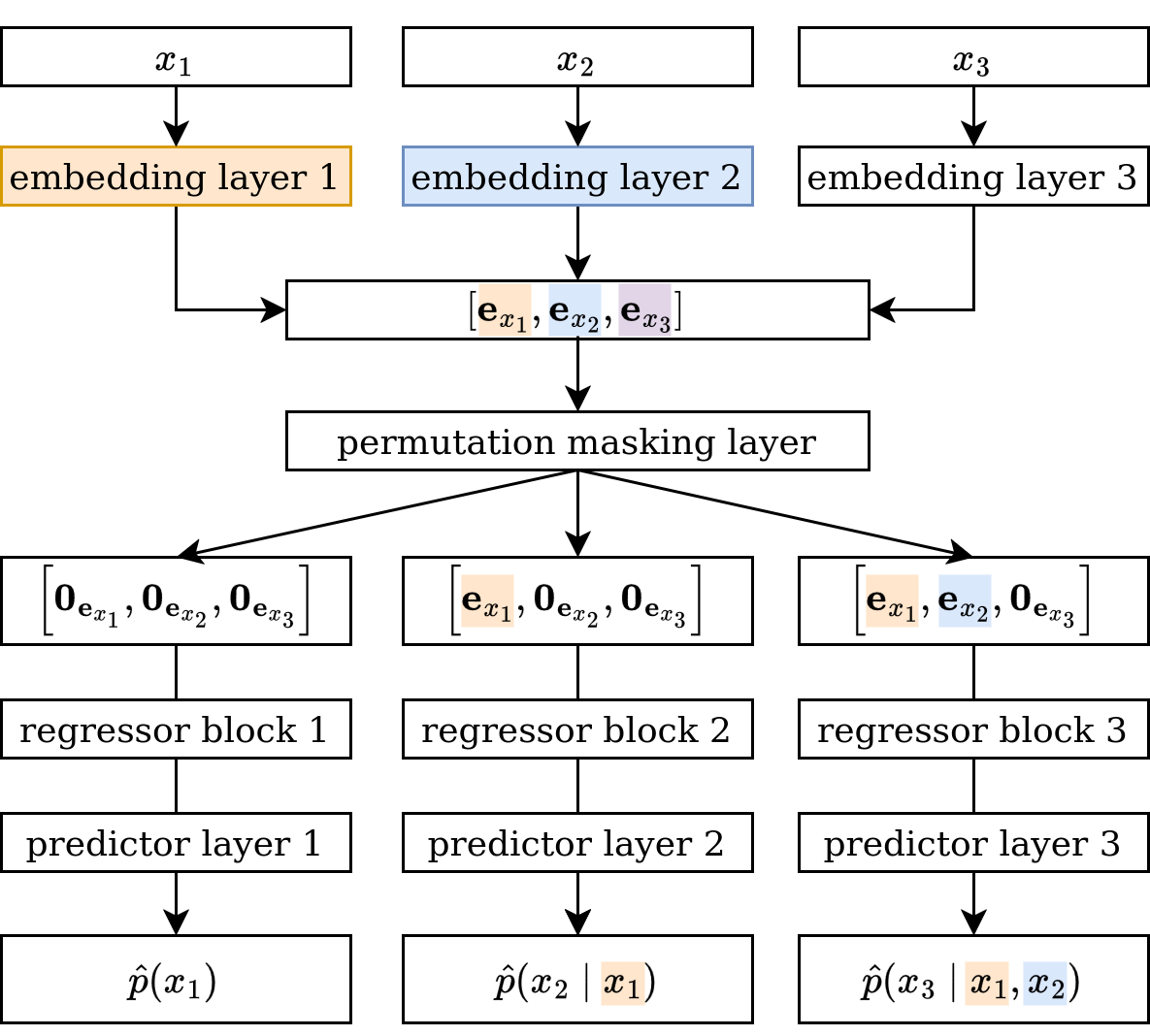}
         \caption{}
         \label{fig:flat-model-train}
     \end{subfigure}
     \hfill
     \begin{subfigure}[b]{0.45\textwidth}
         \centering
         \includegraphics[width=\textwidth]{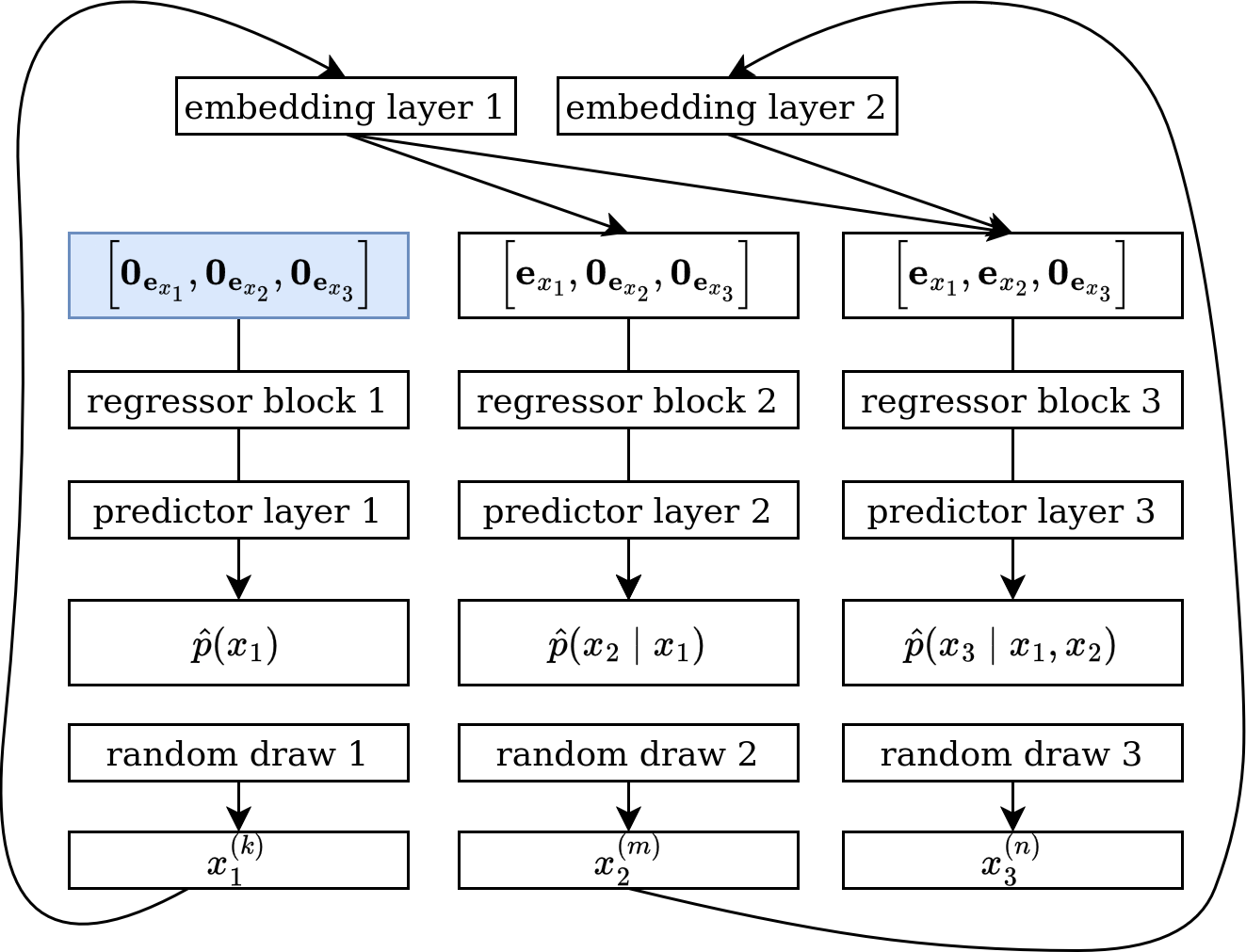}
         \caption{}
         \label{fig:flat-model-gen}
     \end{subfigure}
     \caption{(a) Model components and information flow in the \textbf{training phase} of a three-column TabularARGN flat model with the current column order [1,2,3]. Input features $x_{i}$ are embedded and sent through the permutation masking layer to condition predictions on preceding columns. In the "any-order" training scheme, the permutation masking layer randomly shuffles the column order for each training batch. $\mathbf{0}_{\mathbf{e}_{x_i}}$ denotes a vector of zeros of the same size as $\mathbf{e}_i$, the embedding vector of feature $i$. (b) Model components and information flow in the \textbf{generation phase}. The input to the model and starting point of the generation is a vector of zeros (blue), triggering the successive generation of synthetic features. Due to the permutation of column orders during training, any column order can be realized in the generation phase.}
     % \vskip 0.2in
\end{figure*}

To implement the auto-regressive setup described above, the TabularARGN flat model architecture comprises three main components for each discretized sub-column: an embedding layer, a regressor layer, and a predictor layer. These components are interconnected through a permutation masking layer responsible for the "causal" feeding mechanism based on the current column order (see Figure\ \ref{fig:flat-model-train}).

We begin the explanation at the output stage, moving step-by-step towards the inputs, i.e., embedding layers. At the final stage of the computational path of each sub-column, a predictor layer outputs the estimated conditional probabilities $\hat{p}(x_i \mid x_{<i})$. Each predictor is implemented as a feed-forward layer with a softmax activation function, where the output dimension is equal to the cardinality (number of categories) of the corresponding discretized sub-column.

The input to each predictor layer is provided by a feature-specific regressor layer, which is a feed-forward neural network employing a ReLU activation function. This regressor receives the concatenated embeddings of all sub-columns, processed by the permutation masking layer. The masking layer enforces causality: for a given sub-column $x_i$, it selectively forwards only the embeddings corresponding to the preceding sub-columns while setting to zero the embeddings of subsequent sub-columns, based on the current column order. The masked embeddings thus represent the conditioning information available for the regressor.

To facilitate this masking, each sub-column is first processed by a dedicated embedding layer. This embedding layer maps categorical feature values to embedding vectors. The dimensionality of each embedding vector is dynamically determined based on the cardinality of the corresponding sub-column, allowing embeddings to adapt their complexity according to the feature's characteristics (see Appendix \ref{A:heuristics} for the employed embedding-size heuristic).

The permutation masking layer concatenates these embedding vectors into a single vector and then applies the causal masking. Depending on the current column ordering (fixed or randomly shuffled per batch in the "any-order" training setup), the masking layer ensures that the regressor associated with each sub-column sees embeddings from only its valid preceding sub-columns, with subsequent embeddings masked out (set to zero).

Thus, the full computational path for each sub-column in TabularARGN is as follows:
\begin{itemize}
\item Embedding Layer: Converts each category of the sub-column into a cardinality-dependent embedding vector.
\item Permutation Masking Layer: Concatenates embeddings from all sub-columns and masks them causally according to the current column order.
\item Regressor Layer: Receives the masked embeddings as input and outputs an intermediate representation using a feed-forward network with ReLU activation.
\item Predictor Layer: Receives the intermediate representation from the regressor and outputs the discrete conditional probabilities using a softmax activation.
\end{itemize}

\subsection{Training and Sampling Procedures}
The training target of TabularARGN is the minimization of the categorical cross-entropy, computed and summed up across each sub-column. With the "any-order" permutations, this procedure effectively minimizes the negative log-likelihood
\begin{equation}
    \max_{\theta} \;
    \mathbb{E}_{\sigma}
    \Biggl[
    \sum_{i=1}^{D}
    \log p_{\theta}\bigl(x_{\sigma(i)} \mid x_{\sigma(<i)}\bigr)
    \Biggr],
\end{equation}
where $\sigma$ is a uniformly drawn random permutation of features (sub-columns). $x_{\sigma(<i)}$ denotes all features that precede the feature $i$ in the permutation $\sigma$.

During training, teacher forcing is employed, where ground-truth values from preceding columns are provided as inputs to condition the model. This approach effectively treats the training process as a multi-task problem, wherein each sub-column represents a distinct predictive task.

The training procedure employs a robust early stopping criterion to prevent overfitting (see Section~\ref{sec:privacy-mechanisms}).

In the generation phase (see Figure\ \ref{fig:flat-model-gen}), TabularARGN sequentially generates synthetic data points feature-by-feature based on a selected column order. The permutation masking layer is removed for generation. The process begins with the first feature in the chosen order, whose regressor receives a vector of zeros as input, and the predictor outputs the marginal probability estimate $\hat{p}(x_1)$. A random draw from this discrete distribution determines the synthetic category for this first feature, which is then embedded. This embedding, combined with zeros, is passed to the regressor of the second feature, and the predictor outputs the conditional probability $\hat{p}(x_2 \mid x_1)$. Another random draw determines the synthetic category for the second feature, and the process continues similarly for subsequent features until all synthetic data points have been generated.

\subsection{Privacy Protection Mechanisms in TabularARGN}
\label{sec:privacy-mechanisms}

During the generation phase, synthetic records are produced through probabilistic sampling from trained TabularARGN models, inherently introducing statistical randomness and noise. Furthermore, TabularARGN incorporates several privacy protection mechanisms within its pre-processing and training procedures. These mechanisms are specifically designed to prevent overfitting and encourage generalization, enabling the synthetic data to accurately reflect statistical patterns while significantly reducing the risk of individual data leakage.

The following techniques specifically contribute to robust privacy preservation:

\textbf{Early Stopping}. The training process incorporates an early stopping mechanism to prevent overfitting and ensure efficient convergence. A validation set is separated from the training data, and validation loss is monitored after each epoch. When validation loss stops improving, a patience mechanism reduces the learning rate, eventually halting training if no further improvements are achieved. The model weights of the epoch with the lowest validation loss are retained as the final trained model (for details, see Appendix \ref{A:model-training}).

\textbf{Dropout}. During training, a dropout rate of 25\% is applied to regressor layers. This regularization technique prevents the model from memorizing training data, thereby enhancing generalization and privacy.

\textbf{Value Protection}. Specialized value protection methods for categorical and numerical features further strengthen privacy: \textit{Rare Category Protection} - Infrequent categorical values are replaced with a generalized placeholder token \texttt{\_RARE\_} or values sampled from more common categories, reducing traceability to individuals. The rarity threshold is randomly selected per column, uniformly ranging from 5 to 8. \textit{Extreme Value Protection} - Numeric and date-time outliers are clipped to prevent disclosure of exceptional cases. Values outside the $k$-th highest and lowest thresholds (uniformly chosen between 5 and 8) are adjusted, ensuring no extreme data points compromise privacy.

\textbf{Differential Privacy (DP)}. TabularARGN also supports formal mathematical guarantees of privacy through DP. Differentially-private stochastic gradient descent (DP-SGD) clips and systematically adds calibrated noise to the gradients during the training procedure to limit the impact of individual records while preserving the aggregate statistical properties essential for data utility \cite{TCS-042,10.1145/2976749.2978318,277172}.

\section{Empirical Benchmarking}
To evaluate TabularARGN, we integrate it into the comprehensive benchmarking pipeline released with Continuous Diffusion for Mixed-Type Tabular Data (CDTD) \cite{mueller2025continuousdiffusionmixedtypetabular}. The authors not only present the CDTD model for generating synthetic data but also provide an extensive benchmark along with a complete implementation of their benchmarking pipeline \cite{cdtdrepo}. We integrate TabularARGN into this existing pipeline to situate our method within the context of current state-of-the-art approaches.

The extensive benchmark provided by \citet{mueller2025continuousdiffusionmixedtypetabular} includes a diverse set of synthetic data generation methods: SMOTE \cite{chawla02smote}, ARF \cite{watson23-ARF}, CTGAN \cite{xu19}, TVAE \cite{xu19}, TabDDPM \cite{kot23-TabDDPM}, CoDi \cite{lee23-Codi}, TabSyn \cite{zha24-TabSyn}, and CDTD \cite{mueller2025continuousdiffusionmixedtypetabular}, and applies them to eleven data sets of different sizes and shapes (see Appendix F in Ref.~\cite{mueller2025continuousdiffusionmixedtypetabular}).

To evaluate and compare the performance of these synthetic data generation methods comprehensively, the benchmark employs several metrics:
\begin{itemize}
    \item Statistical Similarity: Evaluated via the Jensen-Shannon divergence (JSD) to measure differences in categorical distributions, the Wasserstein distance (WD) to measure differences in continuous distributions, and the L2 distance between pairwise correlation matrices.
    \item  Detection Score: Assessed by the ability of a classifier (catboost) to distinguish between real and synthetic data samples.
    \item Machine Learning Efficiency: Evaluated following the train-synthetic-test-real approach using logistic/ridge regression, random forest, and catboost models, quantified by RMSE for regression tasks and AUC/macro-averaged F1 scores for classification tasks.
    \item  Privacy Metric (DCR): Measured using DCR, quantifying how closely synthetic records resemble those in the original training data.
\end{itemize}
For more details on the implementation details of the benchmark synthetic-data models and the metrics, we refer the reader to Ref.~\cite{mueller2025continuousdiffusionmixedtypetabular}.

In the remainder of this section, we present the results of our initial benchmarking effort. By "initial," we mean that i) we directly incorporate results from the original CDTD benchmark (Tables 12-19 in Ref.~\cite{mueller2025continuousdiffusionmixedtypetabular}) and supplement them with our results for TabularARGN, without additional modifications or re-evaluation of the original methods; ii) From the list of privacy-protecting mechanisms, we only apply early stopping and dropout during model training, as these techniques represent widely accepted and standard practices in machine learning to reduce overfitting and promote generalization.

We summarize the results in Table \ref{tab:initial-benchmark}, showing the average rank of all evaluated methods across eleven data sets for each metric. To maintain clarity and simplicity, we include only the "CDTD (per type)" variant from the original study, as it consistently represents the strongest performing variant. Following the original benchmark, we \textit{assign the maximum possible rank when a model could not be trained on a given data set or could not be evaluated in a reasonable time.}

TabularARGN performs competitively, achieving two first-place rankings (Wasserstein distance and detection score) and two second-place rankings (L2 distance and AUC) across the seven evaluated metrics. This positions TabularARGN as a robust and competitive method that is on par with current state-of-the-art synthetic data generation methods. The strong performance in the Detection Score is already a pointer to TabularARGN's favorable privacy-utility tradeoff, which we will discuss in the following.

The performance of TabularARGN is also worth noting in light of the model size. In the CDTD benchmark, all models (except SMOTE) are adjusted to a total of ~3 million trainable parameters, which only slightly varies across data sets. As the architecture of TabularARGN models depends on the number of columns, the number of trainable parameters in ARGN fluctuates more, but for most of the data sets, it is a factor of five to ten below the CDTD values (see Table~\ref{tab:model-sizes}). Two notable exceptions are the \texttt{lending} and \texttt{news} data sets with a larger number of numerical columns that are discretized in ARGN into one sub-column per digit.

In general, we want to note that for many data set-metric combinations, performance differences among methods are marginal, underscoring that the comparison represents a very tight competition. Detailed, data set-specific results and a more granular view of performance can be found in Appendix \ref{A:benchmark-results}.

\begin{table*}[ht]
\centering
\begin{tabular}{lccccccccc}
\toprule
metric & smote & arf & ctgan & tvae & tabddpm & codi & tabsyn & cdtd (per type) & ARGN \\
\midrule
RMSE & 4.2 & \underline{3} & 7.0 & 7.4 & 6.2 & 6.8 & 3.8 & \textbf{2.8} & 3.6 \\
F1 score & 3.7 & 5.3 & 6.3 & 6.8 & \underline{3.3} & 5.7 & 7.2 & \textbf{2.7} & 3.5 \\
AUC & 3.7 & 4.7 & 7.3 & 6.7 & 3.2 & 6.0 & 7.0 & \textbf{2.7} & \underline{3.0} \\
L2 distance & 4.0 & 4.8 & 7.2 & 6.8 & 5.1 & 6.0 & 5.9 & \textbf{1.9} & \underline{3.1} \\
Detection score & 4.7 & 5.4 & 7.6 & 6.6 & 4.2 & 6.5 & 5.2 & \underline{2.0} & \textbf{1.8} \\
Jensen-Shannon Divergence & 6.1 & \textbf{1.2} & 6.6 & 7.8 & 5.8 & 6.1 & 5.5 & \underline{2.2} & 3.1 \\
Wasserstein Distance & \underline{2.8} & 5.0 & 6.6 & 7.0 & 4.7 & 7.5 & 5.2 & \underline{2.8} & \textbf{2.4} \\
\bottomrule
\end{tabular}
\caption{Average performance rank of each generative model across eleven data sets. Per metric, bold indicates the best, and underlines the second-best result. The ranking is based on the data from Tables \ref{tab:appendix_auc}-\ref{tab:appendix_WS} in Appendix \ref{A:benchmark-results}, which are, except for the TabularARGN results, extracted from the benchmark in \cite{mueller2025continuousdiffusionmixedtypetabular}. Following the procedure in \cite{mueller2025continuousdiffusionmixedtypetabular}, we assign the highest possible rank if results are missing/impossible to obtain. If two methods score equally - a tie - the lower rank is assigned to both methods.}
\label{tab:initial-benchmark}
\end{table*}

\section{Privacy Evaluation}
Synthetic tabular data can be used for various applications, but a primary use case (and often an implicit expectation) is the protection of individual privacy. This translates into a requirement that synthetic data limits information leakage from the original training data set to the generated synthetic data set.

\subsection{Distance to Closest Record (DCR)}

In the CDTD benchmark, as well as many prior studies, the metric used to quantify this leakage has been DCR, where, e.g., for every synthetic record, the distance to the closest record in the training set is reported \cite{zha24-TabSyn, zha24-TabDar, mueller2025continuousdiffusionmixedtypetabular}. The privacy evaluation involves comparing the distribution of DCR values between training and synthetic data (DCR$_{\text{train/syn}}$) with the distribution between training data and a holdout test set (DCR$_{\text{train/test}}$). Within the DCR framework, synthetic data is considered optimal when the synthetic data is "close" to the training data, ensuring high fidelity, but remains sufficiently distinct to avoid overfitting, indicated by synthetic samples being closer to the training data than test samples are \cite{pla21-qa}.

During the integration of TabularARGN into the CDTD benchmark, we observed that certain methods display mean DCR$_{\text{train/syn}}$ values smaller than the corresponding DCR$_{\text{train/test}}$ (see Table 16 in Ref.~\cite{mueller2025continuousdiffusionmixedtypetabular}). To further investigate this phenomenon, we examine the cumulative distribution functions (CDFs) of the DCR distributions. To accomplish this, we re-ran the training procedures of selected benchmark methods, specifically CDTD, TabDDPM, TabSyn, SMOTE, and ARF, as those are the ones with either strong performance in the quality metrics and/or a low mean DCR$_{\text{train/syn}}$.

\begin{figure}
\includegraphics[width=0.47\textwidth]{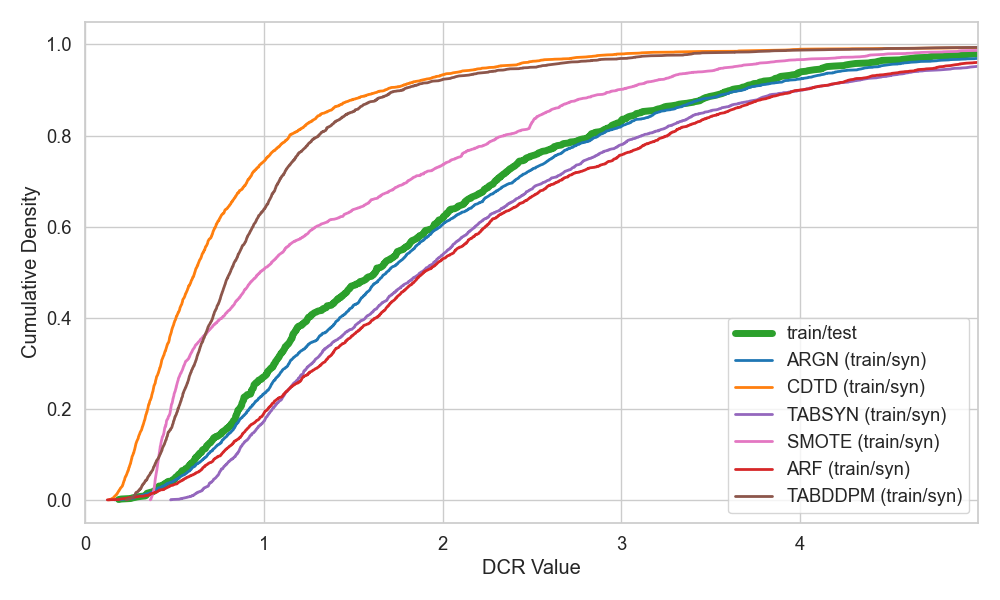}
\centering
\caption{Cumulative density functions of DCR values for \texttt{nmes} data set. Synthetic data generated with CDTD (orange), TABDDPM (brown), and SMOTE (pink) is considerably closer to the training data than a held-out test data set (green), indicating overfitting. TabularARGN (blue) and other methods produce synthetic data that is close to the training data yet slightly more distant than the held-out test data.}
\label{fig:dcr_nmes}
\end{figure}

For the smallest data set in the benchmark (\texttt{nmes}), we observe that the DCR-CDFs for CDTD, TabDDPM, and SMOTE exhibit a clear shift toward smaller distances between synthetic and training samples, signaling significant overfitting (see Figure\ \ref{fig:dcr_nmes}). Other methods show distances slightly larger than the train-test DCRs and CDFs, which roughly follow the shape of the train-test curve. Notably, the overfitting is not limited to the small \texttt{nmes} data set but appears across larger data sets as well (see Figure\ \ref{fig:DCR-grid}).

 The pronounced overfitting observed with SMOTE is anticipated, given its interpolation-based nature. The overfitting in CDTD and TabDDPM is most likely rooted in the lack of an early stopping mechanism and either the lack of or the insufficient use of generalization techniques. TabularARGN uses an early stopping mechanism and dropout rates of 25\% for regressor layers. It seems that both measures ensure that the DCR train-synthetic values of TabularARGN are larger than the DCR train-test values across almost all data sets (see Figure~\ref{fig:DCR-grid}). Also, the shape of the two DCR curves aligns well with the train-test curves.
 
 Given the challenge of defining a straightforward validation loss for diffusion models, we attempted to mitigate overfitting by terminating training early, that is, before the specified number of training steps in the CDTD benchmark. However, this approach frequently resulted in DCR-CDF shapes diverging from the desired distribution (DCR$_{\text{train/test}}$), with areas where synthetic samples were closer to training samples. These observations support our hypothesis that, similar to TabularARGN, additional generalization strategies may be necessary to adequately address this issue. However, we consider such developments outside the scope of this study and propose them for future research.

To systematically account for observed overfitting in our benchmark, we adopt the following strategy: we calculate the integral of the difference between the CDF(DCR$_{\text{train/syn}}$) and the CDF(DCR$_{\text{train/test}}$) from 0 up to the point where the train/test curve reaches 0.98. This upper bound of the integral is chosen for numerical stability and the notion that synthetic records with a large DCR do not pose privacy risks. A negative integral indicates acceptable privacy preservation, whereas a positive integral signifies potential privacy risks. For data set-method combinations yielding positive integrals, we assign the lowest possible rank across all metrics (with appropriate handling of ties: methods with an equal score are assigned the highest possible rank). Detailed integral calculations and results for all data set-method combinations are reported in Table \ref{tab:dcr-integrals}.

Applying this integral-based correction across all data sets and metrics results in a revised ranking (see Table \ref{tab:benchmark-with-privacy}). TabularARGN emerges as the strongest overall method, achieving five first-place rankings and two second-place rankings out of the seven evaluated metrics. Conversely, the CDTD, TabDDPM, and SMOTE methods, demonstrating consistent overfitting behavior, show significantly poorer rankings. In this setting, TabularARGN shows the best trade-off between utility and privacy, within the DCR framework.

\begin{figure*}[!h]
\includegraphics[width=0.8\textwidth]{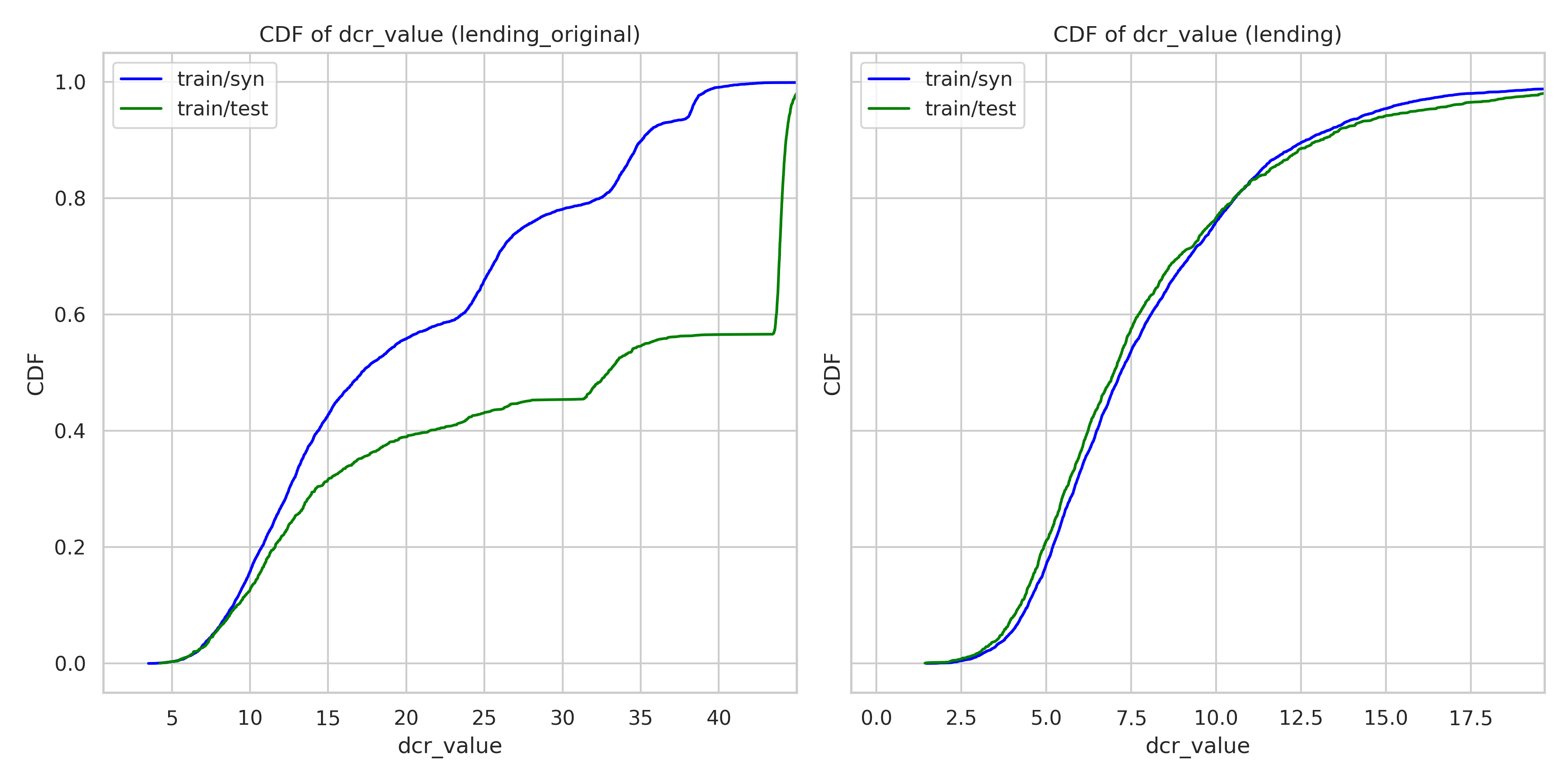}
\centering
\caption{Cumulative density functions of DCR values of TabularARGN (blue) synthetic data and a held-out test set (green) from the \texttt{lending} data set. Without rare-category protection (left), synthetic data is closer to the training data than the holdout, and both CDFs feature steps. With rare-category protection (right) - categories appearing less than eight times are replaced by the \texttt{rare} token during pre-processing - the synthetic data is close to the training data but, for a large fraction of DCR values, slightly more distant than the test set, and both CDFs are smoothed. Switching on the rare-category protection also lowers the DCR values in general. This is most likely caused by reduced distances due to an increase in the matches between categorical columns with a high fraction of \texttt{rare} tokens.}
\label{fig:lending_rare_categories}
\end{figure*}

For the \texttt{lending} data set, we observe that all methods, including TabularARGN, have train-synthetic DCR distributions considerably shifted to the left compared to the train-test curves (see bottom left plot in Figure\ \ref{fig:DCR-grid}. This implies that, at least for specific data sets, early stopping and dropout are not enough to avoid synthetic data being closer to training data than a test set is. In the case of the \texttt{lending} data set, the issue is caused by high-cardinality columns, specifically column \texttt{emp\_title}, which has almost 5k unique values across the 10k rows. Almost 98\% of categories appear less than eight times. We mitigate this effect by protecting these rare categories through collecting them in a new "catch-all" category during pre-processing. We scan through all categorical columns and overwrite every category that appears less than eight times across the data set with the token \texttt{rare}\footnote{This approach is very similar to TabularARGN's rare-category protection. We decided to use a constant cut-off of eight for reproducibility.}. The effect on the DCR CDFs is significant (see Figure\ \ref{fig:lending_rare_categories}). Not only do both CDFs (train-syn and train-test) become smoother, but the TabularARGN DCRs transform from being closer to the training data to being close, but slightly larger than the train-test DCRs.

At this stage, it remains unclear whether the observed "shift-to-the-left" in the synthetic data results from overfitting or from a potential bias in the generative models to produce more frequent samples from denser regions of the original distribution. We leave the investigation and detailed discussion of this phenomenon for future research.

Beyond the limited interpretability of the DCR CDFs, we acknowledge that the DCR metric has its limitations in reliably measuring privacy due to its reliance on distance-based measures, which may not capture subtler aspects of information leakage \cite{yao2025dcrdelusionmeasuringprivacy}. Despite these limitations, DCR is frequently utilized in practice due to its computational simplicity, practicality, and capability of detecting generative overfitting to the training data. To rigorously evaluate TabularARGN's privacy performance, thoroughly acknowledge the DCR shortcomings, and investigate further the effects of rare category protection, we dedicate the following sections to a comprehensive analysis of the specific privacy-protecting mechanisms implemented in TabularARGN and subject the method to more robust privacy evaluations through membership-inference attacks.

\begin{table*}[ht]
\centering
\begin{tabular}{lccccccccc}
\toprule
metric & SMOTE & ARF & CTGAN & TVAE & TABDDPM & CODI & TabSyn & CDTD (per type) & ARGN \\
\midrule
RMSE & 6.6 & \textbf{2.2} & 5.0 & 5.0 & 5.8 & 4.2 & 2.6 & 5.2 & \underline{2.4} \\
F1 score & 7.3 & \underline{3.2} & 4.7 & 4.7 & 5.0 & 3.8 & 5.0 & 6.5 & \textbf{2.0} \\
AUC & 7.3 & \underline{2.5} & 5.2 & 4.5 & 5.0 & 4.2 & 5.0 & 6.7 & \textbf{2.0} \\
L2 distance & 7.0 & \underline{3.1} & 5.3 & 4.7 & 5.5 & 3.7 & 3.7 & 5.9 & \textbf{1.9} \\
Detection score & 7.0 & \underline{3.3} & 5.5 & 4.5 & 5.2 & 4.1 & \underline{3.3} & 5.9 & \textbf{1.5} \\
Jensen-Shannon Divergence & 7.0 & \textbf{1.2} & 4.7 & 5.5 & 5.9 & 4.1 & 4.0 & 5.9 & \underline{2.3} \\
Wasserstein Distance & 7.0 & \underline{3.2} & 4.8 & 4.8 & 5.2 & 5.1 & \underline{3.2} & 5.8 & \textbf{1.5} \\
\bottomrule
\end{tabular}
\caption{Average performance rank of each generative model across eleven data sets similar to Table \ref{tab:initial-benchmark}. In this ranking, we consider potential information leakage by assigning the highest possible rank to a method across all metrics if the CDF-DCR integral $\int_0^{0.98}CDF(DCR_{\text{train/syn}}) - CDF(DCR_{\text{train/test}})$ is positive (Table \ref{tab:dcr-integrals}), i.e., indicates overfitting. If two methods score equally - a tie - the lower rank is assigned to both methods.}
\label{tab:benchmark-with-privacy}
\end{table*}

\subsection{Membership-Inference Attacks (MIAs)}

To thoroughly evaluate privacy robustness, we conduct systematic Membership-Inference Attacks (MIAs) on synthetic data sets generated with TabularARGN under different configurations: i) without protection beyond dropout and early stopping, ii) with value protection enabled, and iii) with DP. The analysis focuses on the \texttt{Adult} data set, a widely used benchmark in privacy research, using the same sample of 1,000 records as in Refs.~\cite{10.1007/978-3-031-51476-0_19,yao2025dcrdelusionmeasuringprivacy}. Two records, identified by their unique or atypical attribute combinations, are selected for detailed evaluation based on high vulnerability scores computed via the Achilles procedure \cite{10.1007/978-3-031-51476-0_19}. This method quantifies a record’s risk by calculating the mean cosine distance to its $k$ nearest neighbors ($k=5$), incorporating both categorical and continuous features through one-hot encoding and min-max normalization, respectively. Records with larger average distances are presumed to be more distinct within the data set and thus more susceptible to inference attacks.

All attack experiments are conducted under a black-box threat model \cite{hayes2018loganmembershipinferenceattacks,wu2024inferenceattackstaxonomysurvey}. In this setting, the adversary is assumed to have full knowledge of the generative model’s structure and output size but no access to its internal parameters or training dynamics. Moreover, the attacker is presumed to possess the entire training data set and the target record itself. This aligns with standard formulations in MIA research and ensures that any inference success results from the information revealed through the synthetic data alone.

To simulate the generative process and train attack models, we employ the shadow modeling technique, a widely accepted method for membership inference evaluation \cite{wu2024inferenceattackstaxonomysurvey,277172}. Specifically, 1,000 shadow models are trained on auxiliary data sets drawn from the same underlying distribution as the original data but excluding the target record. Half of these shadow data sets included the target record, while the other half substituted it with a randomly chosen alternative. Each shadow model is then used to produce a synthetic data set, labeled according to the presence or absence of the target record in the training data. These labeled synthetic data sets serve as training data for a meta-classifier, which was subsequently applied to synthetic data sets generated from the true training set to infer the membership of the target records.

To evaluate a broad range of membership inference strategies, we rely on a modified version of TAPAS (Toolbox for Adversarial Privacy Auditing of Synthetic Data) \cite{houssiau2022tapastoolboxadversarialprivacy}. We extended TAPAS to support benchmarking with TabularARGN, allowing its full suite of implemented attacks to be applied directly to the synthetic datasets generated by our framework. These attack strategies include classical shadow model-based attacks, such as the Groundhog suite (Naive, Histogram-based, Correlation-based, and Logistic regression-based variants), which operate by learning aggregate statistical features from the synthetic data (for details, see Appendix \ref{A:attacks}). More advanced shadow modeling attacks are also tested, including the extended TAPAS variant that takes advantage of query-based features derived from counting how often subsets of the attributes of a target record appear in the synthetic data sets \cite{10.1007/978-3-031-51476-0_19,yao2025dcrdelusionmeasuringprivacy}. These query-based attacks are particularly effective when subtle distributional shifts occur due to the inclusion of the target record.

For the sake of completeness, we perform distance-based inference attacks that evaluate record similarity using Hamming and Euclidean (L2) metrics. These attacks assign higher membership scores to synthetic records closely matching the target record. A direct lookup strategy is also used to test whether the exact target record appeared in the synthetic data. Finally, we evaluate inference-on-synthetic attacks using a kernel density estimator, which models the synthetic data distribution and infers membership based on the likelihood of the target record under this distribution \cite{houssiau2022tapastoolboxadversarialprivacy,277172,steier2025syntheticdataprivacymetrics}.

Attack success is measured using classification accuracy and the area under the receiver operating characteristic curve (AUC). Accuracy reflects the proportion of correct member/non-member predictions, while AUC provides a threshold-independent measure of discriminability. An AUC of 0.5 indicates random guessing, whereas values significantly above 0.5 reflect successful inference and, hence, privacy leakage.

This methodological framework allows us to assess the privacy risks posed by synthetic data generation comprehensively, including evaluating the impact of value protection and differential privacy on MIA performance. By focusing on records identified as the most vulnerable through the Achilles scoring method, we ensure that the evaluation targets the worst-case privacy risks inherent in synthetic data release. This design supports a robust understanding of both the capabilities of different MIA strategies and the protective efficacy of existing mitigation mechanisms.

\subsection{MIAs Results}

\begin{figure*}[!h]
\includegraphics[width=\textwidth]{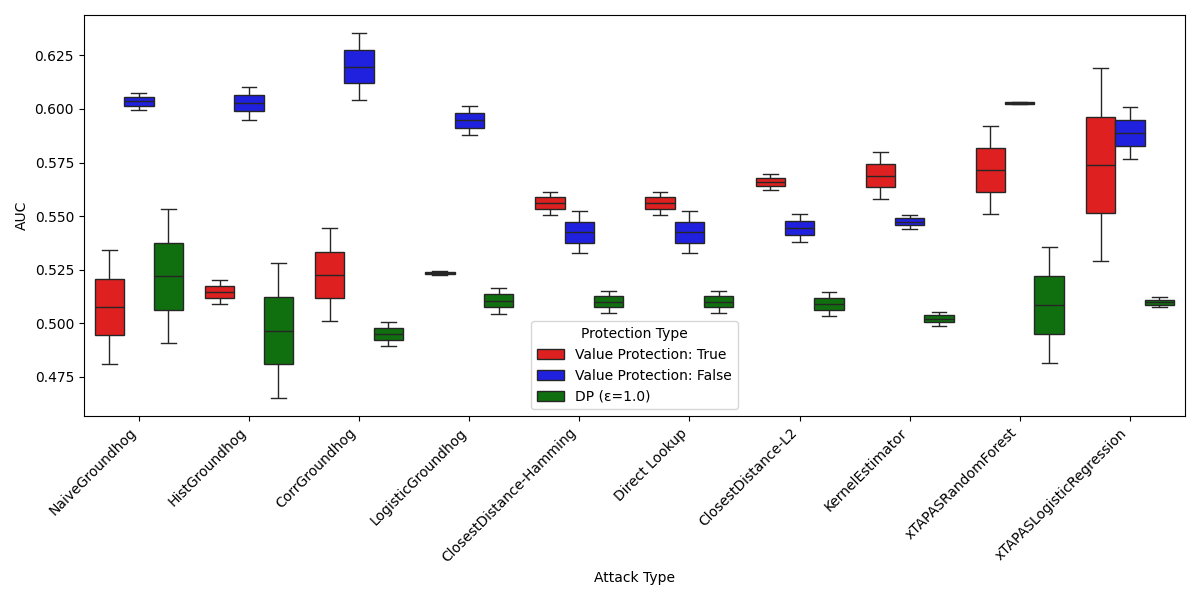}
\centering
\caption{Area Under the Curve (AUC) of membership inference attacks on two most vulnerable target records in the \texttt{Adult} data set grouped by protection type  - without both value protection and DP (blue), with only value protection (red), and with the DP budget $\epsilon=1.0$ (green). The distinction clearly demonstrates the impact of protection mechanisms and DP settings on the MIA's AUC, with visible variations across attack types.}
\label{fig:auc_by_attack}
\end{figure*}

The analysis of MIAs performed on the two most vulnerable records from the \texttt{Adult} data set without DP and special protection of rare values reveals limited but subtle privacy leakage. As can be seen in Figure~\ref{fig:auc_by_attack} (blue boxes), although attack methods demonstrate moderate effectiveness, the majority of AUC values are only slightly above the baseline of 0.5, ranging from approximately 0.48 to 0.64. This indicates that while synthetic data does not flagrantly expose records, it retains subtle privacy vulnerabilities.

The highest observed effectiveness is achieved by the Correlation Groundhog attack based on pairwise attribute correlations,\cite{277172}, which yields an AUC of $0.635$ and an accuracy of about $60.1\%$ in scenarios without the rare value protection and DP. This means that an attacker can correctly identify members about $60\%$ of the time versus $50\%$ by chance, indicating moderate leakage of membership information. Other advanced attacks (e.g., Histogram Groundhog, Logistic Groundhog, or extended-TAPAS attacks) also hover around $57-59\%$ accuracy ($\text{AUC}\approx0.58-0.60$). The marginal gains observed in extended-TAPAS attacks compared to simpler heuristics indicate that for certain data characteristics, ensemble methods may not significantly outperform simpler approaches. This aligns with existing literature \cite{houssiau2022tapastoolboxadversarialprivacy}, emphasizing that while advanced methods possess potential, their superiority is context-dependent.

In addition, even simpler threshold-based attacks like Naive Groundhog also succeed above chance. The Naive Groundhog in this case achieves $\sim\!57\%$ accuracy ($\text{AUC}\approx0.61$) without protections. This is strong evidence that model confidence is correlated with membership, i.e., training samples tend to yield higher confidence predictions of the correct class than unfamiliar samples, enabling a naive confidence-threshold attack to outperform random guessing.

On the other hand, the distance-based scores are significantly lower ($\text{AUC}\approx0.53-0.56$ for Hamming, L2, and Direct lookup), which means the generator does not produce exact or extremely close replicas of this record, i.e., the leakage is more subtle, through distributional effects rather than verbatim memorization.

Record-specific analysis reveals differential vulnerabilities between the two Achilles-identified records. One record consistently demonstrates slightly higher vulnerability across most attack methods and scenarios, indicating inherent record-specific characteristics that enhance susceptibility to privacy leakage. This aligns with \citet{10.1007/978-3-031-51476-0_19}, who note that records with unique or outlying attributes are generally more vulnerable, as the synthetic generator may struggle to effectively obscure such records.

When the value protection mechanism is enabled, it significantly influences the attack effectiveness, reducing privacy leakage. The attack AUC values generally decrease to the random guessing level of $0.50-0.55$, indicating effective mitigation against straightforward and sophisticated attacks alike (see orange boxes in Figure~\ref{fig:auc_by_attack}). Specifically, CorrGroundhog’s AUC drops notably from 0.635 without protection to approximately 0.501 with the protection, highlighting the substantial efficacy of the protection mechanism against privacy risks associated with unique or rare attribute values.

\begin{figure*}[!h]
\includegraphics[width=\textwidth]{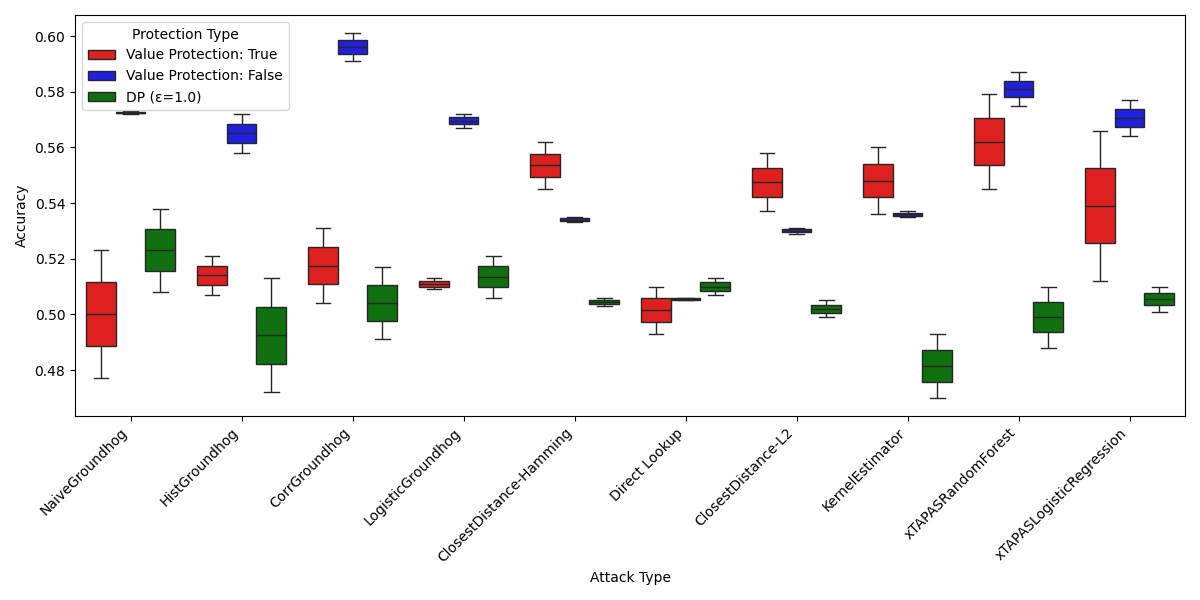}
\centering
\caption{Accuracy of membership inference attacks on the two most vulnerable target records in the \texttt{Adult} data set grouped by protection type  - without both value protection and DP (blue), with only value protection (red), and with the DP budget $\epsilon=1.0$ (green). The distinction clearly demonstrates the impact of protection mechanisms and DP settings on the MIA accuracy, with visible variations across attack types.}
\label{fig:accuracy_by_attack}
\end{figure*}

Additionally, we evaluate the effect of applying DP to the synthetic generator. With DP ($\epsilon=1.0$ in our experiment), all attacks collapse to near-chance performance. For both records, AUC plummets to $\sim\!0.50-0.52$ and attack accuracy to $\sim\!50-51\%$, essentially nullifying the membership advantage (see Figure~\ref{fig:auc_by_attack}). This is an encouraging sign that a sufficiently strong DP guarantee can protect even the most vulnerable individuals. Our results mirror the findings of \citet{10.1007/978-3-031-51476-0_19}, that as the privacy budget $\epsilon$ decreases, MIA success rates drop significantly. In fact, in our case, $\epsilon=1.0$ is enough to reduce the MIA AUC to 0.5. (We note that $\epsilon\approx1.0$ is a fairly strict privacy level). Of course, strong DP often comes at the cost of synthetic data utility. Nonetheless, from a purely privacy perspective, our empirical evidence is that differential privacy is highly effective at thwarting these membership attacks, essentially randomizing away the signals that attackers are exploiting.

\section{Discussion}

While achieving a high utility in the benchmarks, the vulnerability levels of TabularARGN synthetic data observed in the baseline scenario, despite being notable, are substantially lower than those frequently reported in other research involving different synthetic data generation methods. For example, studies utilizing GANs and related advanced techniques typically report higher AUC values (often exceeding 0.7) \cite{277172,10.1007/978-3-031-51476-0_19,yao2025dcrdelusionmeasuringprivacy}. This comparison highlights the inherent privacy-preserving strengths of the TabularARGN, even in the absence of explicit protective measures. We link these lower vulnerabilities to the early stopping mechanism and the large fraction of dropout applied during the training of TabularARGN.

Applying the value protection mechanism dramatically reduces this vulnerability even further. Once enabled, the performance of attacks dropped to near-random levels. For both records, the AUC plummets to roughly $0.5-0.6$, and the attack accuracy falls to almost $50\%$, essentially equivalent to guessing. This marks a substantial improvement from the baseline, indicating that the value protection method effectively neutralizes the unique patterns or outlier values that the attack leverages. In practical terms, after targeted perturbations, the adversary can no longer confidently distinguish the protected records as training members. This result highlights the power of focused, record-specific defense, i.e., by identifying and sanitizing rare values in the data set, the methodology removes the telltale artifacts that enable membership inference. Notably, this approach is in line with recent recommendations to address privacy risks by handling the most vulnerable data points individually. For example, removing or modifying outlier records is known to significantly decrease overall membership inference risk \cite{10.5555/3600270.3601234}. The value protection embodies this principle, yielding a pronounced privacy gain for minimal change in the data. Furthermore, because the protection is applied in a fine-grained manner (at the level of specific values or records), the technique can preserve much of the data set’s utility while still thwarting the attack on those sensitive entries. Utility is typically preserved because the value protection affects mainly very rare categories which only contribute a very low signal in, e.g, downstream machine learning tasks on synthetic data. Our findings show that a well-chosen heuristic defense can, in this case, be as effective as far more complex strategies in safeguarding vulnerable records.

Introducing DP-SGD into the training (with privacy budget $\epsilon=1.0$) similarly curtailed the attacker’s advantage. Under DP protection, the membership inference AUC for the two target records drops to near-chance (approximately 0.55 or lower), and the attack’s accuracy hovers close to $50\%$. This substantial reduction in attack efficacy indicates that DP’s noise injection successfully obfuscates the membership information. Intuitively, the stochastic perturbations required by $\epsilon=1.0$ prevent the model from memorizing fine-grained details of individual records, thereby suppressing the very signals that the extended TAPAS attack was exploiting. These results are consistent with established findings that DP can provide a robust defense against MIAs by adding uncertainty to model outputs. \citet{chen2021differential} for example, observe that a differentially private model produces noisy prediction vectors such that an adversary \textit{cannot easily infer the membership} of any given sample. 
In our experiments, even a moderate privacy budget ($\epsilon=1$) is enough to appreciably blunt the attack, underscoring DP’s effectiveness. However, as expected, this protection comes with trade-offs. The DP-trained model likely experienced some loss in accuracy or fidelity due to the noise, a well-known consequence of DP \cite{ijcai2021p432}. Prior work cautions that while DP is a powerful defense, it often imposes a non-trivial utility cost, especially for complex models or high-dimensional data. Thus, the DP scenario in our study demonstrates a substantial privacy improvement, but one must weigh the benefit against potential reductions in the overall performance of the model or the quality of synthetic data.

Comparing the value protection and DP side-by-side reveals interesting findings. Both defenses significantly lower the membership inference AUC and accuracy, effectively protecting the test records, yet their mechanisms and implications differ. The value protection achieves nearly the same reduction in attack success as DP, despite not providing a formal privacy guarantee. This suggests that much of the baseline risk is concentrated in a few identifiable features of the records, which the value protection mechanism successfully obscures. A key advantage of this approach is its precision, i.e., by only altering the values of the known vulnerable record(s), it preserves model utility on the rest of the data. In contrast, DP offers worst-case privacy guarantees covering all records, but at the cost of perturbing the training process globally. From a utility perspective, the model with the value protection approach likely retains higher accuracy on benign inputs than the DP model, since DP’s random noise perturbs many parameters indiscriminately. In practice, a combination of strategies might even be considered. For instance, using the value protection to safeguard particularly sensitive records on top of a mild DP baseline to harness the strengths of each. The broader implication is that defenses need not be "one-size-fits-all": a strategic, point-specific intervention can sometimes rival the protection of a uniform, theoretically grounded defense (DP) for the cases that matter most.

\section{Conclusion}

The current work introduces TabularARGN and analyzes its inherent strengths: a favorable privacy-utility tradeoff. It provides high utility and a robust baseline privacy compared to other synthetic data generation methods reported elsewhere. Our experiments suggest that this effect is due to the early stopping and the considerable dropout rates applied during the training of TabularARGN. Enhanced privacy measures, such as value protection and differential privacy, further mitigate vulnerabilities effectively. In practice, organizations must weigh these factors. Our results suggest that when a generative model shows clear “Achilles’ heels” in its training data, addressing those with minimalistic interventions can dramatically improve privacy. On the other hand, if one requires formal assurances for all data points, approaches like DP (despite their impact on accuracy) become attractive. Ultimately, the results show that both targeted and general defenses substantially mitigate membership inference risk for the tested records, indicating that strong privacy protection is achievable, either through pinpoint fixes or through robust noise-based frameworks, without completely sacrificing utility. By carefully selecting a defense strategy appropriate to the threat model and data characteristics, one can significantly reduce the chance that an attacker successfully infers membership, thus bolstering the privacy of individuals in the data set.

That said, the present work also leaves several important questions open that should be addressed in future research. For instance, it remains to be explored how the benchmark models used in this study perform under standard regularization techniques, such as early stopping and dropout. If DCR scores no longer indicate overfitting under these conditions, a direct comparison of synthetic data quality between these models and TabularARGN would provide a more balanced utility perspective. Furthermore, a rigorous evaluation of utility and synthetic data quality when applying additional privacy protection mechanisms, such as value protection and DP-SGD, would deepen our understanding of the tradeoffs involved. We leave these investigations to future work and hope they will inspire further progress in privacy-preserving synthetic data generation.

\begin{acks}
We extend our deepest gratitude to Michael Platzer, Klaudius Kalcher, Roland Boubela, Felix Dorrek, and Thomas Gamauf, whose foundational ideas and early contributions were instrumental in shaping many of the key architectural components of TabularARGN and guiding its initial implementation.

We are also grateful to the following colleagues for their substantial and wide-ranging contributions: Ivona Krchova, Mariana Vargas Vieyra, Mario Scriminacci, Radu Rogojanu, Lukasz Kolodziejczyk, Michael Druk, Shuang Wu, Andr\'e Jonasson, Dmitry Aminev, Peter Bognar, Victoria Labmayr, Manuel Pasieka, Daniel Soukup, Anastasios Tsourtis, Jo\~ao Vidigal, Kenan Agyel, Bruno Almeida, Jan Valendin, and M{\"u}rsel Ta\c{s}g{\i}n. Their collective efforts played a vital role in refining TabularARGN, enhancing its performance, robustness, and usability while ensuring strong safeguards against information leakage.

A special thanks to Ivona Krchova for her careful proofreading and invaluable feedback, which helped improve the clarity and quality of this work.
\end{acks}

\bibliographystyle{ACM-Reference-Format}
\bibliography{biblio}

\appendix

\section{Heuristics, Layer, and Model Sizes}
\label{A:heuristics}

The table below provides an overview of the layer sizes and dimensions for the embedding, regressor, and predictor layers.

\begin{table}[h]
\centering
\begin{tabular}{lc}
Layer & Heuristic \\
\toprule
Embeddings & $3 \cdot (d_{\text{in}})^{\mathstrut 0.25} $ \\
Regressors & $16 \cdot \text{max}(1, \text{ln}(d_{\text{in}}))$ \\
Predictors & $d_{\text{in}}$ \\
\bottomrule
\end{tabular}
\caption{Heuristic for calculating the size of the model layers for a sub-column with cardinality $d_{\text{in}}$.}
\label{tab:heuristics}
\end{table}

For the data sets used in this benchmark, this results in the following number of trainable parameters in TabularARGN:

\begin{table}[h]
\centering
\begin{tabular}{lrrr}
\hline
Dataset & \makecell{Number of\\Parameters} & \makecell{Categorical\\Features} & \makecell{Continuous\\Features} \\
\hline
acsincome & 220729 & 8 & 3 \\
adult & 289316 & 9 & 6 \\
bank & 619406 & 11 & 10 \\
beijing & 302531 & 1 & 10 \\
churn & 349918 & 5 & 9 \\
covertype & 483969 & 44 & 10 \\
default & 996198 & 2 & 14 \\
diabetes & 785678 & 28 & 9 \\
lending & 4335217 & 10 & 34 \\
news & 11736069 & 14 & 46 \\
nmes & 452106 & 8 & 11 \\
\hline
\end{tabular}
\caption{Number of trainable parameters of the TabularARGN model across data sets. The number of features/columns influences the model size.}
\label{tab:model-sizes}
\end{table}

\section{Model Training}
\label{A:model-training}

To ensure efficient and robust model training, TabularARGN employs an early stopping mechanism based on validation loss. This mechanism prevents overfitting and reduces unnecessary training time by halting the process when further improvements in validation performance are unlikely.

Validation Loss and Early Stopping: During training, 10\% of the input data set is split off as a validation set. At the end of each epoch, the model’s validation loss is calculated. If the validation loss does not improve for $N$ consecutive epochs, training is stopped. The default setting for $N$ is 5 epochs. The model weights corresponding to the lowest observed validation loss are retained as the final trained model.

Learning Rate Scheduler: In addition to early stopping, a learning rate scheduler is employed to dynamically adjust the learning rate during training. If the validation loss does not improve for $K$ consecutive epochs, the learning rate is halved to promote finer adjustments in the model's parameters. The default setting for $K$ is 3 epochs.

These mechanisms work together to optimize the training process, ensuring that the model converges efficiently while minimizing the risk of overfitting.

\section{Detailed Benchmark results}
\label{A:benchmark-results}

Here we present detailed benchmarking results across model types and data sets organized by metric. Each table shows the corresponding results from Ref.~\cite{mueller2025continuousdiffusionmixedtypetabular} and supplements these with the new performance metrics for TabularARGN.  As stated in \cite{mueller2025continuousdiffusionmixedtypetabular}: \textit{CoDi is prohibitively expensive to train on \texttt{lending} and \texttt{diabetes}, and} TabDDPM \textit{produces} NaNs \textit{for \texttt{acsincome} and \texttt{diabetes}.} SMOTE \textit{takes too long to sample data sets of a sufficient size for \texttt{acsincome} and \texttt{covertype}}. Additionally, we extend the original benchmark by providing plots of the cumulative distribution functions (CDFs) of the DCR distributions (Figure \ref{tab:dcr-integrals}) and include a table (see \ref{tab:dcr-integrals}) reporting the integral of the difference between the CDF(DCR$_{\text{train/syn}}$) and the CDF(DCR$_{\text{train/test}}$).

\begin{table*}[ht]
\centering
\begin{tabular}{lrrrrrrrrrrr}
\toprule
data set & SMOTE & ARF & CTGAN & TVAE & TabDDPM & CoDi & TabSyn & \makecell{Tabular\\ARGN} & \makecell{CDTD\\(single)} & \makecell{CDTD\\(per type)} & \makecell{CDTD\\(per feature)} \\
\midrule
acsincome & NaN & 0.242 & 1.696 & 1.136 & NaN & 0.517 & 0.560 & 0.168 & 0.226 & 0.166 & 0.163 \\
adult & 0.414 & 0.576 & 1.858 & 0.735 & 0.160 & 0.493 & 0.514 & 0.182 & 0.215 & 0.153 & 0.216 \\
bank & 0.404 & 0.819 & 0.947 & 2.758 & 0.529 & 0.499 & 0.759 & 0.222 & 0.600 & 0.456 & 0.542 \\
beijing & 0.081 & 0.128 & 1.470 & 1.226 & 0.368 & 0.373 & 0.086 & 0.134 & 0.170 & 0.192 & 0.095 \\
churn & 0.264 & 0.635 & 1.355 & 1.301 & 0.273 & 0.746 & 0.613 & 0.701 & 0.528 & 0.476 & 0.456 \\
covertype & NaN & 1.192 & 3.685 & 4.668 & 1.124 & 1.029 & 3.749 & 0.786 & 2.107 & 1.366 & 1.880 \\
default & 0.709 & 1.228 & 2.697 & 1.564 & 0.685 & 1.672 & 1.125 & 0.568 & 1.370 & 1.153 & 1.058 \\
news & 1.684 & 4.333 & 4.641 & 4.612 & 12.356 & 4.874 & 5.153 & 4.014 & 4.372 & 4.025 & 4.385 \\
nmes & 0.565 & 0.717 & 1.663 & 0.532 & 0.426 & 0.609 & 0.919 & 0.711 & 3.506 & 3.506 & 0.834 \\
diabetes & 2.355 & 1.189 & 1.654 & 5.351 & NaN & NaN & 2.796 & 0.731 & 1.381 & 1.213 & 1.351 \\
lending & 1.321 & 3.473 & 2.420 & 5.895 & 10.046 & NaN & 6.792 & 3.629 & 1.148 & 1.239 & 1.351 \\
\bottomrule
\end{tabular}
\caption{$L_2$ norm of the correlation matrix differences of synthetic data and original training data for seven benchmark models, CDTD with three different noise schedules, and TabularARGN.}
\label{tab:appendix_l2}
\end{table*}

\begin{table*}[ht]
\centering
\begin{tabular}{lrrrrrrrrrrr}
\toprule
data set & SMOTE & ARF & CTGAN & TVAE & TabDDPM & CoDi & TabSyn & \makecell{Tabular\\ARGN} & \makecell{CDTD\\(single)} & \makecell{CDTD\\(per type)} & \makecell{CDTD\\(per feature)} \\
\midrule
acsincome & NaN & 0.013 & 0.256 & 0.309 & NaN & 0.076 & 0.052 & 0.024 & 0.064 & 0.050 & 0.053 \\
adult & 0.064 & 0.007 & 0.112 & 0.113 & 0.035 & 0.045 & 0.022 & 0.021 & 0.031 & 0.028 & 0.031 \\
bank & 0.039 & 0.004 & 0.086 & 0.191 & 0.021 & 0.038 & 0.063 & 0.010 & 0.035 & 0.036 & 0.032 \\
beijing & 0.006 & 0.004 & 0.005 & 0.074 & 0.024 & 0.011 & 0.011 & 0.006 & 0.031 & 0.025 & 0.031 \\
churn & 0.012 & 0.011 & 0.095 & 0.048 & 0.015 & 0.043 & 0.031 & 0.014 & 0.034 & 0.031 & 0.033 \\
covertype & NaN & 0.002 & 0.044 & 0.043 & 0.004 & 0.008 & 0.044 & 0.003 & 0.010 & 0.005 & 0.007 \\
default & 0.042 & 0.008 & 0.194 & 0.177 & 0.028 & 0.073 & 0.093 & 0.018 & 0.049 & 0.040 & 0.036 \\
news & 0.063 & 0.002 & 0.022 & 0.128 & 0.017 & 0.012 & 0.017 & 0.006 & 0.017 & 0.022 & 0.020 \\
nmes & 0.060 & 0.008 & 0.117 & 0.029 & 0.025 & 0.027 & 0.019 & 0.014 & 0.380 & 0.380 & 0.030 \\
diabetes & 0.067 & 0.009 & 0.093 & 0.187 & NaN & NaN & 0.098 & 0.020 & 0.024 & 0.025 & 0.031 \\
lending & 0.143 & 0.049 & 0.092 & 0.188 & 0.287 & NaN & 0.119 & 0.110 & 0.056 & 0.057 & 0.065 \\
\bottomrule
\end{tabular}
\caption{Jensen-Shannon divergence for seven benchmark models, CDTD with three different noise schedules, and TabularARGN.}
\label{tab:appendix_JSD}
\end{table*}

\begin{table*}[ht]
\centering
\begin{tabular}{lrrrrrrrrrrr}
\toprule
data set & SMOTE & ARF & CTGAN & TVAE & TabDDPM & CoDi & TabSyn & \makecell{Tabular\\ARGN} & \makecell{CDTD\\(single)} & \makecell{CDTD\\(per type)} & \makecell{CDTD\\(per feature)} \\
\midrule
acsincome & NaN & 0.007 & 0.037 & 0.021 & NaN & 0.017 & 0.005 & 0.002 & 0.008 & 0.002 & 0.002 \\
adult & 0.003 & 0.012 & 0.016 & 0.021 & 0.002 & 0.013 & 0.007 & 0.003 & 0.008 & 0.003 & 0.004 \\
bank & 0.002 & 0.012 & 0.021 & 0.040 & 0.004 & 0.030 & 0.006 & 0.002 & 0.012 & 0.012 & 0.014 \\
beijing & 0.002 & 0.008 & 0.030 & 0.036 & 0.007 & 0.019 & 0.004 & 0.003 & 0.019 & 0.018 & 0.004 \\
churn & 0.006 & 0.013 & 0.027 & 0.032 & 0.007 & 0.048 & 0.013 & 0.010 & 0.014 & 0.019 & 0.012 \\
covertype & NaN & 0.006 & 0.041 & 0.022 & 0.002 & 0.012 & 0.019 & 0.003 & 0.020 & 0.008 & 0.013 \\
default & 0.002 & 0.005 & 0.011 & 0.005 & 0.002 & 0.013 & 0.003 & 0.003 & 0.009 & 0.006 & 0.005 \\
news & 0.007 & 0.024 & 0.009 & 0.018 & 0.033 & 0.030 & 0.029 & 0.006 & 0.026 & 0.023 & 0.026 \\
nmes & 0.005 & 0.012 & 0.036 & 0.008 & 0.007 & 0.016 & 0.032 & 0.005 & 0.048 & 0.048 & 0.041 \\
diabetes & 0.004 & 0.012 & 0.020 & 0.038 & NaN & NaN & 0.012 & 0.004 & 0.042 & 0.034 & 0.041 \\
lending & 0.006 & 0.013 & 0.011 & 0.016 & 0.410 & NaN & 0.053 & 0.012 & 0.012 & 0.013 & 0.011 \\
\bottomrule
\end{tabular}
\caption{Wasserstein Distance for seven benchmark models, CDTD with three different noise schedules, and TabularARGN.}
\label{tab:appendix_WS}
\end{table*}

\begin{table*}[ht]
\centering
\begin{tabular}{lrrrrrrrrrrr}
\toprule
data set & SMOTE & ARF & CTGAN & TVAE & TabDDPM & CoDi & TabSyn & \makecell{Tabular\\ARGN} & \makecell{CDTD\\(single)} & \makecell{CDTD\\(per type)} & \makecell{CDTD\\(per feature)} \\
\midrule
acsincome & NaN & 0.808 & 0.989 & 0.985 & NaN & 0.825 & 0.688 & 0.535 & 0.611 & 0.573 & 0.572 \\
adult & 0.687 & 0.889 & 0.997 & 0.967 & 0.594 & 0.992 & 0.641 & 0.560 & 0.641 & 0.608 & 0.609 \\
bank & 0.839 & 0.955 & 1.000 & 0.988 & 0.781 & 1.000 & 0.853 & 0.546 & 0.823 & 0.833 & 0.850 \\
beijing & 0.938 & 0.989 & 0.996 & 0.995 & 0.738 & 0.989 & 0.723 & 0.678 & 0.777 & 0.766 & 0.651 \\
churn & 0.567 & 0.853 & 0.945 & 0.843 & 0.556 & 0.730 & 0.859 & 0.820 & 0.787 & 0.827 & 0.736 \\
covertype & NaN & 0.945 & 0.997 & 0.989 & 0.584 & 0.900 & 0.991 & 0.809 & 0.991 & 0.973 & 0.986 \\
default & 0.928 & 0.991 & 0.998 & 0.997 & 0.827 & 0.995 & 0.914 & 0.763 & 0.901 & 0.852 & 0.902 \\
news & 0.998 & 0.998 & 1.000 & 1.000 & 0.974 & 1.000 & 0.999 & 0.921 & 0.994 & 0.995 & 0.996 \\
nmes & 0.926 & 0.987 & 0.992 & 0.988 & 0.652 & 0.988 & 0.829 & 0.578 & 0.984 & 0.983 & 0.823 \\
diabetes & 0.726 & 0.854 & 0.935 & 0.997 & NaN & NaN & 0.946 & 0.613 & 0.864 & 0.837 & 0.862 \\
lending & 0.966 & 0.997 & 0.995 & 0.995 & 1.000 & NaN & 0.998 & 0.977 & 0.959 & 0.955 & 0.960 \\
\bottomrule
\end{tabular}
\caption{Detection score for seven benchmark models, CDTD with three different noise schedules, and TabularARGN.}
\label{tab:appendix_detection}
\end{table*}

\begin{table*}[ht]
\centering
\begin{tabular}{lrrrrrrrrrrrr}
\toprule
data set & SMOTE & ARF & CTGAN & TVAE & TabDDPM & CoDi & TabSyn & Test Set & \makecell{Tabular\\ARGN} & \makecell{CDTD\\(single)} & \makecell{CDTD\\(per type)} & \makecell{CDTD\\(per feature)} \\
\midrule
acsincome & NaN & 8.637 & 10.758 & 6.652 & NaN & 10.877 & 10.797 & 7.673 & 8.882 & 9.345 & 9.199 & 9.188 \\
adult & 1.371 & 2.523 & 5.012 & 2.227 & 1.656 & 2.735 & 2.408 & 1.870 & 2.262 & 1.465 & 1.557 & 1.646 \\
bank & 1.369 & 3.025 & 3.840 & 3.136 & 2.211 & 3.062 & 3.022 & 2.369 & 2.645 & 2.536 & 2.383 & 2.522 \\
beijing & 0.139 & 0.731 & 0.800 & 0.724 & 0.639 & 0.588 & 0.633 & 0.385 & 0.598 & 0.657 & 0.662 & 0.534 \\
churn & 0.232 & 1.136 & 1.804 & 1.146 & 0.368 & 0.852 & 1.209 & 0.347 & 1.099 & 0.797 & 0.853 & 0.688 \\
covertype & NaN & 1.741 & 5.773 & 3.173 & 0.877 & 1.508 & 3.033 & 0.529 & 1.026 & 2.274 & 1.524 & 1.780 \\
default & 1.032 & 3.095 & 5.880 & 3.216 & 1.437 & 2.593 & 2.801 & 1.812 & 2.042 & 1.900 & 1.851 & 1.839 \\
news & 3.553 & 6.147 & 4.789 & 5.821 & 4.358 & 4.661 & 5.410 & 3.615 & 4.297 & 4.638 & 4.550 & 4.683 \\
nmes & 1.394 & 2.203 & 2.971 & 1.710 & 0.890 & 1.231 & 2.105 & 1.931 & 2.039 & 3.185 & 3.185 & 1.979 \\
diabetes & 13.909 & 17.736 & 21.935 & 8.214 & NaN & NaN & 28.794 & 15.608 & 17.103 & 14.356 & 14.468 & 14.866 \\
lending & 17.752 & 17.776 & 20.239 & 10.688 & 14.310 & NaN & 16.239 & 11.184 & 24.255 & 14.958 & 14.962 & 14.146 \\
\bottomrule
\end{tabular}
\caption{Mean distance to closest record (DCR) for seven benchmark models, CDTD with three different noise schedules, and TabularARGN.}
\label{tab:appendix_dcr}
\end{table*}

\begin{table*}[ht]
\centering
\begin{tabular}{lrrrrrrrrrrrr}
\toprule
data set & SMOTE & ARF & CTGAN & TVAE & TabDDPM & CoDi & TabSyn & Training Set & \makecell{Tabular\\ARGN} & \makecell{CDTD\\(single)} & \makecell{CDTD\\(per type)} & \makecell{CDTD\\(per feature)} \\
\midrule
adult & 0.784 & 0.769 & 0.647 & 0.756 & 0.788 & 0.745 & 0.782 & 0.797 & 0.796 & 0.788 & 0.792 & 0.780 \\
bank & 0.740 & 0.682 & 0.680 & 0.629 & 0.744 & 0.673 & 0.661 & 0.745 & 0.718 & 0.737 & 0.720 & 0.714 \\
churn & 0.865 & 0.780 & 0.761 & 0.802 & 0.855 & 0.865 & 0.748 & 0.873 & 0.784 & 0.819 & 0.838 & 0.849 \\
covertype & NaN & 0.783 & 0.442 & 0.711 & 0.799 & 0.767 & 0.620 & 0.817 & 0.781 & 0.743 & 0.768 & 0.766 \\
default & 0.677 & 0.627 & 0.686 & 0.632 & 0.680 & 0.638 & 0.485 & 0.674 & 0.676 & 0.673 & 0.675 & 0.659 \\
diabetes & 0.615 & 0.572 & 0.557 & 0.553 & NaN & NaN & 0.566 & 0.621 & 0.604 & 0.614 & 0.619 & 0.614 \\
\bottomrule
\end{tabular}
\caption{Machine learning efficiency F1 score for seven benchmark models, CDTD with three different noise schedules, TabularARGN, and the original training data.}
\label{tab:appendix_f1}
\end{table*}

\begin{table*}[ht]
\centering
\begin{tabular}{lrrrrrrrrrrrr}
\toprule
data set & SMOTE & ARF & CTGAN & TVAE & TabDDPM & CoDi & TabSyn & Training Set & ARGN & \makecell{CDTD\\(single)} & \makecell{CDTD\\(per type)} & \makecell{CDTD\\(per feature)} \\
\midrule
adult & 0.906 & 0.901 & 0.836 & 0.889 & 0.909 & 0.880 & 0.906 & 0.915 & 0.912 & 0.907 & 0.908 & 0.904 \\
bank & 0.943 & 0.938 & 0.934 & 0.830 & 0.942 & 0.929 & 0.922 & 0.947 & 0.943 & 0.940 & 0.940 & 0.939 \\
churn & 0.961 & 0.939 & 0.882 & 0.948 & 0.957 & 0.961 & 0.911 & 0.964 & 0.925 & 0.949 & 0.955 & 0.955 \\
covertype & NaN & 0.860 & 0.677 & 0.777 & 0.876 & 0.845 & 0.675 & 0.892 & 0.858 & 0.825 & 0.847 & 0.841 \\
default & 0.759 & 0.754 & 0.744 & 0.751 & 0.765 & 0.739 & 0.732 & 0.768 & 0.761 & 0.762 & 0.765 & 0.761 \\
diabetes & 0.679 & 0.669 & 0.626 & 0.592 & NaN & NaN & 0.642 & 0.693 & 0.678 & 0.671 & 0.673 & 0.672 \\
\bottomrule
\end{tabular}
\caption{Machine learning efficiency AUC score for seven benchmark models, CDTD with three different noise schedules, TabularARGN, and the original training data.}
\label{tab:appendix_auc}
\end{table*}

\begin{table*}[ht]
\centering
\begin{tabular}{lrrrrrrrrrrrr}
\toprule
data set & SMOTE & ARF & CTGAN & TVAE & TabDDPM & CoDi & TabSyn & Training Set & ARGN & \makecell{CDTD\\(single)} & \makecell{CDTD\\(per type)} & \makecell{CDTD\\(per feature)} \\
\midrule
acsincome & NaN & 0.757 & 2.292 & 1.054 & NaN & 0.857 & 0.955 & 0.804 & 0.842 & 0.894 & 0.834 & 0.851 \\
beijing & 0.742 & 0.779 & 1.050 & 1.295 & 0.799 & 0.849 & 0.789 & 0.711 & 0.758 & 0.794 & 0.818 & 0.772 \\
news & 1.180 & 0.923 & 1.906 & 3.999 & 0.171 & 1.302 & 0.397 & 1.001 & 1.098 & 0.571 & 0.533 & 0.588 \\
nmes & 1.112 & 0.972 & 1.331 & 1.127 & 1.200 & 1.137 & 0.563 & 1.001 & 1.098 & 0.417 & 0.417 & 1.162 \\
lending & 0.042 & 0.274 & 0.137 & 0.404 & 0.789 & NaN & 0.305 & 0.030 & 0.275 & 0.071 & 0.066 & 0.062 \\
\bottomrule
\end{tabular}
\caption{Machine learning efficiency RMSE for seven benchmark models, CDTD with three different noise schedules, TabularARGN, and the original training data.}
\label{tab:appendix_rmse}
\end{table*}

\begin{figure*}[!h]
\includegraphics[width=0.89\textwidth]{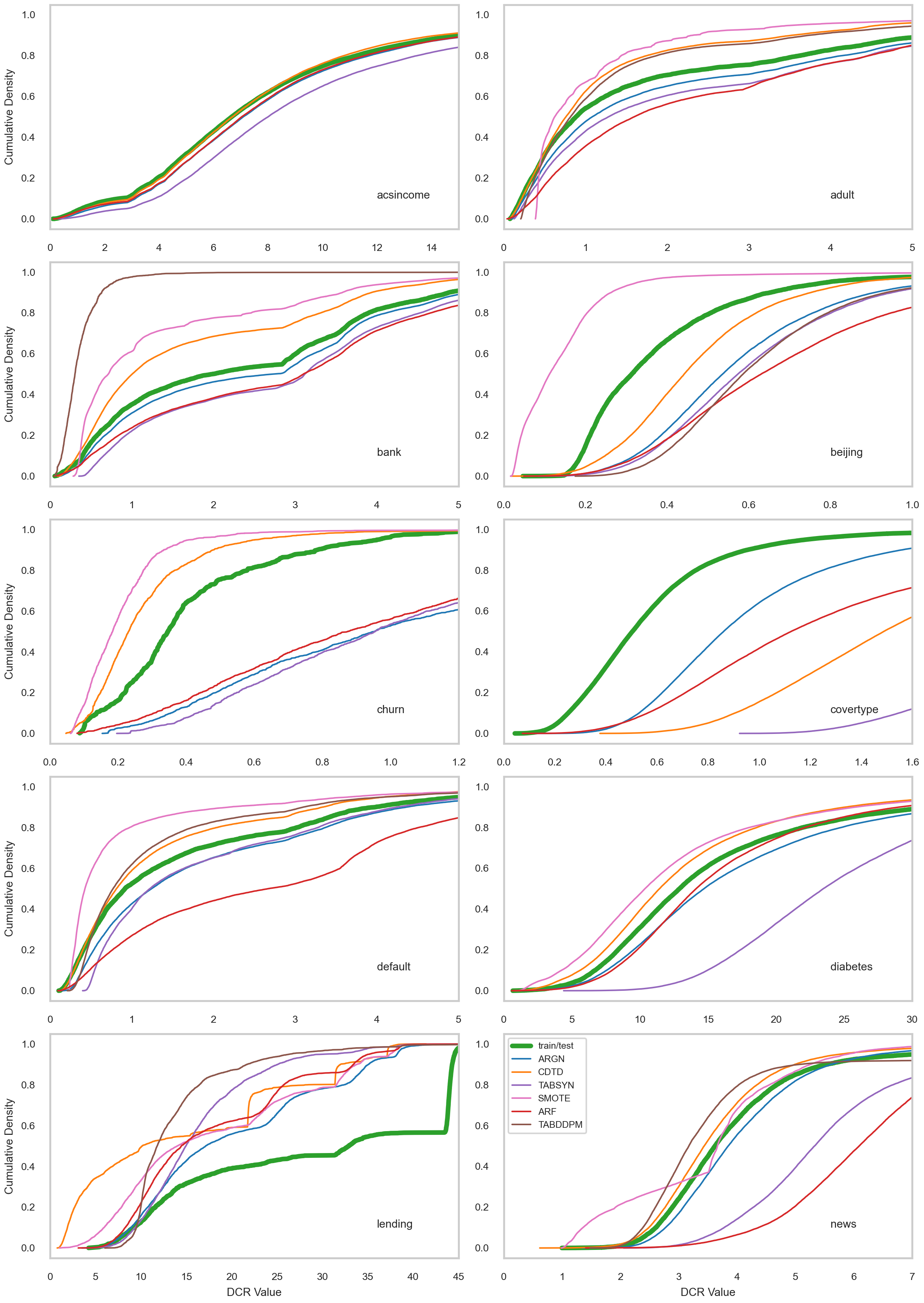}
\centering
\caption{Cumulative densities of the DCR distributions $\text{DCR}_{\text{train/test}}$ (fat green) and $\text{DCR}_{\text{train/syn}}$ from TabularARGN (blue), CDTD (per type) (orange), TabSyn (violet), SMOTE (pink), ARF (red), and TabDDPM (brown) across data sets.}
\label{fig:DCR-grid}
\end{figure*}

\begin{table*}[ht]
\centering
\begin{tabular}{lrrrrrr}
\toprule
data set & TabularARGN & CDTD & TabSyn & SMOTE & ARF & TabDDPM \\
\midrule
acsincome & -0.46 & 0.14 & -1.89 & NaN & -0.34 & NaN \\
adult & -0.29 & 0.62 & -0.54 & 0.74 & -0.59 & 0.43 \\
bank & -0.20 & 0.66 & -0.57 & 1.01 & -0.61 & 2.01 \\
beijing & -0.21 & -0.10 & -0.24 & 0.24 & -0.31 & -0.26 \\
churn & -0.59 & 0.14 & -0.59 & 0.20 & -0.53 & NaN \\
covertype & -0.39 & -0.95 & -1.45 & NaN & -0.68 & NaN \\
default & -0.29 & 0.32 & -0.31 & 0.73 & -1.19 & 0.35 \\
diabetes & -1.60 & 3.08 & -8.08 & 3.74 & -0.01 & NaN \\
lending & 8.80 & 14.19 & 12.68 & 11.13 & 11.06 & 15.13 \\
news & -0.15 & 0.29 & -1.60 & 0.44 & -2.19 & 0.27 \\
nmes & -0.09 & 1.03 & -0.29 & 0.46 & -0.30 & 0.86 \\
\bottomrule
\end{tabular}
\caption{Values of the integral $\int_0^{0.98}CDF(DCR_{\text{train/syn}}) - CDF(DCR_{\text{train/test}})$ across different generative models and data sets. The upper bound of 0.98 for the integral is selected for numerical stability and the notion that synthetic records with large DCRs do not pose a privacy risk.}
\label{tab:dcr-integrals}
\end{table*}

\section{Attack methodologies}
\label{A:attacks}
Attack methodologies, each leveraging different principles to infer membership:

\begin{itemize}
    \item \textit{Groundhog Variants} (Naive Groundhog, Hist Groundhog, Corr Groundhog, Logistic Groundhog): These attacks are rooted in the shadow modeling technique, as conceptualized by~\citet{277172}. Their operation involves training a meta-classifier on aggregate statistics derived from synthetic shadow data sets. Stadler et al. specifically utilized features such as the mean and standard deviation of attributes, correlation matrices, and histograms. This directly informs the naming conventions observed, such as "Hist Groundhog" and "Corr Groundhog." "Naive Groundhog" and "Logistic Groundhog" likely refer to the specific type of meta-classifier (e.g., logistic regression) or simpler feature sets employed within this framework.

    \item \textit{Distance-Based Methods} (Closest Distance-Hamming, Closest Distance-L2, Direct Lookup): These methods fall under the category of "Local Neighborhood Attacks" within the TAPAS framework \cite{houssiau2022tapastoolboxadversarialprivacy}. Their core principle is to assess membership based on the proximity of the target record to records within the synthetic data set. The attack score is typically inversely proportional to the distance to the closest synthetic record. "Direct Lookup" is a specialized instance, effectively checking for an exact match (distance = 0). "Closest Distance-Hamming" is particularly suited for categorical data, while "Closest Distance-L2" (Euclidean distance) is designed for continuous data. The Achilles paper itself utilizes a generalized cosine distance that combines both attribute types.

    \item \textit{Kernel Estimator}: This attack likely represents an "Inference-on-Synthetic" approach. In this methodology, a density model, such as a Kernel Density Estimator, is fitted directly to the synthetic data. The attack then assesses the likelihood or density of the target record under this learned distribution, with higher likelihood potentially indicating membership.

    \item \textit{extended-TAPAS Variants} (\textit{extended-TAPAS with Random Forest} and \textit{extended-TAPAS with Logistic Regression}): These are advanced shadow-modeling attacks, specifically implementations of the "query-based attack" \cite{10.1007/978-3-031-51476-0_19,yao2025dcrdelusionmeasuringprivacy}. They train a meta-classifier (e.g., Random Forest or Logistic Regression, as indicated by their names) on "counting queries" ($Q^{A}$). These queries determine the frequency of synthetic records matching specific subsets of attribute values of the target record. The underlying principle is that the presence of a target record in the training data is likely to locally influence the synthetic data generation, leading to statistically different answers to these queries.
\end{itemize}

\end{document}